\documentclass[12pt, a4paper]{article}

\usepackage[utf8]{inputenc}
\usepackage[T1]{fontenc}
\usepackage[english]{babel}

\usepackage{geometry}
\usepackage{graphicx}
\usepackage{float}
\usepackage{booktabs}
\usepackage{subcaption}
\usepackage{tabularx}

\usepackage{amsmath}
\usepackage{amsfonts}
\usepackage{amssymb}

\usepackage[ruled,vlined,linesnumbered]{algorithm2e}

\usepackage{listings}
\usepackage{xcolor}

\definecolor{codegreen}{rgb}{0,0.6,0}
\definecolor{codegray}{rgb}{0.5,0.5,0.5}
\definecolor{codepurple}{rgb}{0.58,0,0.82}
\definecolor{backcolour}{rgb}{0.95,0.95,0.92}
\definecolor{magenta}{rgb}{1.0, 0.0, 1.0}

\lstdefinestyle{mystyle}{
    backgroundcolor=\color{backcolour},
    commentstyle=\color{codegreen},
    keywordstyle=\color{magenta},
    numberstyle=\tiny\color{codegray},
    stringstyle=\color{codepurple},
    basicstyle=\ttfamily\footnotesize,
    breakatwhitespace=false,
    breaklines=true,
    captionpos=b,
    keepspaces=true,
    numbers=left,
    numbersep=5pt,
    showspaces=false,
    showstringspaces=false,
    showtabs=false,
    tabsize=2,
    literate={ą}{{\k{a}}}1 {ć}{{\'c}}1 {ę}{{\k{e}}}1 {ł}{{\l}}1 {ń}{{\'n}}1 {ó}{{\'o}}1 {ś}{{\'s}}1 {ż}{{\.z}}1 {ź}{{\'z}}1 {Ą}{{\k{A}}}1 {Ć}{{\'C}}1 {Ę}{{\k{E}}}1 {Ł}{{\L}}1 {Ń}{{\'N}}1 {Ó}{{\'O}}1 {Ś}{{\'S}}1 {Ż}{{\.Z}}1 {Ź}{{\'Z}}1
}
\usepackage{url}
\usepackage[hidelinks]{hyperref}

\title{\Large\textbf{The K-SCAN Clustering Algorithm}}
\author{}
\date{}
\begin{document}

\maketitle

\vspace{0.55cm}

\centerline{\large Filip Kosiorowski} \vspace{4mm}
\centerline{\footnotesize\it  The Faculty of Mathematics and Applied
Physics, Rzesz\'ow University of Technology,}
\centerline{\footnotesize\it Powsta\'nc\'ow Warszawy 12, 35-959 Rzesz\'ow, Poland}
\centerline{\footnotesize\it e-mail: kosiorowskifilip@gmail.com}
\vspace{0.25cm}
\centerline{\large Grzegorz Sroka } \vspace{4mm}
\centerline{\footnotesize\it Department of Analysis Nonlinear, Rzesz\'ow University of Technology,}
\centerline{\footnotesize\it Powsta\'nc\'ow Warszawy 12, 35-959 Rzesz\'ow, Poland}
\centerline{\footnotesize\it e-mail: gsroka@prz.edu.pl}
\vspace{0.4cm}

% ============================================================
% ABSTRACT
% ============================================================
\noindent\textbf{Abstract.} In the Big Data era, the scalability of clustering algorithms constitutes a key challenge. Traditional density-based methods (e.g., DBSCAN) offer robustness to noise and the ability to detect non-linear clusters, yet their quadratic time complexity $O(N^2)$ drastically limits their applicability. Conversely, partitional algorithms (e.g., K-Means), with their linear complexity $O(N)$, impose sphericity on the resulting groups and fail in the presence of outliers. This paper presents K-SCAN -- a novel hybrid algorithm that optimizes this trade-off. The method integrates preliminary vector quantization (stochastic Mini-Batch K-Means) to extract a reduced set of weighted micro-clusters, followed by a subsequent density-based structural analysis. Empirical evaluation on datasets of up to $10^6$ samples confirms the linear computational complexity of the proposed solution. K-SCAN achieves more than a 3-fold speed-up over the hierarchical BIRCH algorithm, avoiding the costly management of tree-based structures. The method precisely identifies non-linear manifolds while maintaining structural stability (Adjusted Rand Index > 0.99), even with noise levels reaching 55\% of the data volume. The main limitation of the proposed algorithm, which could not be fully eliminated in the present study, remains its susceptibility to over-smoothing and its difficulty in separating clusters with highly heterogeneous local density. In complex visual spaces, this can lead to the loss of the finest topological details.

\vspace{0.5cm}

\section{Introduction}

The rapid growth of data volume in the Big Data era drives the need for highly scalable unsupervised learning methods \cite{bishop2006pattern}, with particular emphasis on cluster analysis \cite{jain1999data}. However, the choice of an appropriate strategy for partitioning the input space involves significant methodological trade-offs. The classic partitional algorithm K-Means \cite{lloyd1982least, macqueen1967some} is characterized by a desirable, linear computational complexity $O(N)$, yet its effectiveness is limited by the a priori assumption of spherical cluster shapes and its high sensitivity to noise and centroid initialization.

Density-based approaches, on the other hand, represented by the DBSCAN algorithm \cite{ester1996density}, effectively isolate outliers and identify groups of arbitrary geometry; however, their quadratic time complexity $O(N^2)$ constitutes a barrier that prevents the processing of large datasets. A response to this dilemma is offered by two-stage hybrid algorithms, which align with modern trends toward reducing the problem space before the actual structural segmentation \cite{tsai2010hybrid, wierzchon2018modern}.

This paper presents the original K-SCAN algorithm, whose overarching goal is to bridge the technological gap between the efficiency of partitional methods and the geometric flexibility of density-based approaches. The proposed solution employs a micro-clustering technique for data compression while preserving key information about the topology of the feature space. The paper puts forward the research thesis that applying preliminary vector quantization via the stochastic Mini-Batch K-Means algorithm \cite{sculley2010web} makes it possible to reduce the computational complexity of density-based clustering to a linear level, while retaining the ability to detect non-spherical structures and noise robustness comparable to that of the classic DBSCAN algorithm \cite{ester1996density, schubert2017dbscan}.

The scientific argument of the paper follows a clear narrative structure. Following the introduction, Section 2 provides a concise theoretical review of the baseline algorithms that form the foundation of the hybrid. Section 3 contains a detailed description of the K-SCAN architecture, a formal pseudocode specification, and a computational complexity analysis. The results of internal experiments, focused on examining the stability of the control parameters, the influence of data dimensionality, and the degree of input-space compression, are presented in Section 4. Section 5 is devoted to a multi-criteria external validation and comparative studies against established reference methods on large, noisy datasets. The paper concludes with Section 6, which synthesizes the results obtained, critically assesses the limitations of the method, and outlines promising directions for its further development.
% ============================================================
% Related Works
% ============================================================
\section{Related Work}

Classical approaches to cluster analysis rely on two dominant paradigms: partitional and density-based \cite{jain_taxonomy, xu2005survey}. The K-Means algorithm \cite{lloyd1982least}, thanks to its low, linear computational complexity, has for decades remained the most widely used method. However, it has fundamental limitations: it imposes spherical cluster shapes and exhibits high sensitivity to outliers (noise) \cite{jain2010data, han2011data}. An alternative is offered by density-based algorithms such as DBSCAN \cite{ester1996density}, which precisely identify clusters of arbitrary geometry and isolate noise. Unfortunately, their pessimistic $O(N^2)$ time complexity constitutes a serious bottleneck when processing Big Data volumes \cite{schubert2017dbscan, berkhin2006survey}.

The development of large-scale analytical systems has driven the creation of data-reduction mechanisms applied prior to the actual clustering step. A classic solution in this area is the BIRCH algorithm \cite{zhang1996birch}, which, in a single pass over the database, builds a hierarchical tree of micro-clusters (CF-Tree), substantially compressing the search space. In the field of partitional methods, in turn, a breakthrough came with Sculley's introduction of the Mini-Batch K-Means algorithm \cite{sculley2010web}. The use of Stochastic Gradient Descent (SGD) together with batch sampling made it possible to drastically shorten the centroid-update time without significant quality degradation, and it is now the industry standard for quantization.

The most dynamically developing trend today is that of hybrid clustering algorithms. Combining the speed of partitional methods with the topological precision of density-based models was already postulated by Tsai and Wu \cite{tsai2010hybrid} and by Wierzchoń \cite{wierzchon2018modern}. More recent work continues in this direction: fusing K-Means \cite{lloyd1982least} with Gaussian Mixture Models (GMM) \cite{reynolds2009gaussian} through the averaging of cluster labels has been shown to stabilize the analysis of data with heterogeneous covariance, and other efforts pursue parameter-free hybrids that combine grid mapping with density-based analysis to support fully autonomous, scalable clustering in noisy environments.

A separate aspect, strongly emphasized in the most recent literature (2025--2026), is the explainability of unsupervised learning algorithms (XAI). Research on textual and graph data analysis shows that classical methods, such as spectral clustering \cite{ng2002spectral}, can gain interpretability through the application of rough set theory \cite{pawlak1982rough}. Starosta and Wierzchoń, in their recent work, likewise emphasize that algorithms must not only process data quickly but also deliver interpretable results \cite{starosta2025rough, starosta2026explainable, klopotek2026applicability}. The proposed K-SCAN algorithm draws on the idea of hybridization, focusing directly on optimizing computational overhead -- it replaces the costly tree- and matrix-based analyses of the extraction phase with fast Mini-Batch quantization, thereby answering the contemporary demand for scalable, noise-resistant Big Data mining methods.
% ============================================================
% BASELINE ALGORITHMS
% ============================================================
\section{Theoretical Foundations of the Analyzed Algorithms}

Cluster analysis \cite{wierzchon2018modern} is a fundamental data mining problem that consists in partitioning a set of objects into subsets such that elements within a given group exhibit maximal similarity, while simultaneously maximizing the distinctness between different groups. Classically, this task is solved using two leading paradigms: the partitional and the density-based.

The main representative of partitional methods is the K-Means algorithm, which performs the task of dividing the space into a predefined number $k$ of disjoint clusters. This process proceeds through the iterative minimization of a quantization error, defined as the sum of the squared Euclidean distances between points and their assigned centroids. The mechanics of the method rely \cite{macqueen1967some} on alternately performing two operations: assigning each point to its nearest centroid, which geometrically corresponds to a partition of the space into Voronoi cells \cite{aurenhammer1991voronoi}, and recomputing the centroid coordinates as the arithmetic means of their assigned points.

Although the classic Lloyd heuristic \cite{lloyd1982least} is characterized by a favorable, linear computational complexity $O(N)$, its deterministic nature forces the assignment of every point to some cluster, which, combined with its tendency to approximate only spherical structures, makes it extremely sensitive to the presence of outliers.

In response to the lack of scalability of the full K-Means algorithm for Big Data volumes, a stochastic variant known as Mini-Batch K-Means was developed \cite{sculley2010web}. This method fundamentally changes the processing structure -- instead of optimizing over the entire dataset at every iteration, it employs stochastic gradient descent operating on small, randomly sampled batches of data. The centroid update is incremental in nature, with the influence of new points scaled by a dynamically decreasing learning rate that depends on the number of samples already assigned. This solution drastically reduces the cost of a single algorithmic step, enabling deep compression of large-scale data spaces in a fraction of the time required by the classical version.

A fundamentally different construction is presented by the DBSCAN algorithm \cite{ester1996density}, which forms the core of the density-based paradigm. Rather than operating on global centroids, DBSCAN analyzes local topological relationships, defining a cluster as a contiguous region of high point density, separated from other structures by sparse zones. Its mechanics rely on verifying two global parameters: the radius of the neighborhood and the minimum required point mass. Identifying a point that satisfies the density criterion triggers a recursive cluster-expansion process (flooding), which absorbs all spatially connected dense regions. Points that fail to satisfy this condition and lie outside the reach of dense structures are permanently isolated as noise. The limitation of this flexible architecture is the need to compute distances within the full neighborhood of every point, which generates a pessimistic computational complexity on the order of $O(N^2)$, constituting a barrier for large-scale applications.

Comparing the discussed algorithms in terms of their mechanics, structure, and design, one can observe a fundamental conflict between speed and topological precision. Algorithms from the K-Means family (including Mini-Batch) \cite{lloyd1982least, sculley2010web} have a flat, matrix-based structure built around the optimization of a global objective function. Their mechanics rely on aggressively compressing the space into a finite number of representatives, which guarantees high efficiency but geometrically forces convexity (sphericity) and leads to the destructive absorption of noise by the centroids. The Mini-Batch variant \cite{sculley2010web} improves only the iterative layer (batch processing), inheriting all the topological shortcomings of the original.

DBSCAN \cite{ester1996density}, in turn, relies on a graph-based neighborhood-search structure and local-expansion mechanics. Such a design frees the model from the requirement of specifying the number of clusters a priori and allows precise mapping of non-linear manifolds and rejection of background (noise); however, the high cost of maintaining spatial relationships causes its performance to drop drastically in the Big Data regime. This contrast constitutes a direct motivation for the design of hybrid systems.

% ============================================================
% DEVELOPMENT OF THE PROPOSED ALGORITHM
% ============================================================
\section{Development of the Proposed K-SCAN Hybrid Algorithm}

This chapter presents a detailed description of the concept, architecture, and implementation of the proposed K-SCAN algorithm. It includes a theoretical analysis of the method, a formal pseudocode specification, a visualization of the algorithm's operating mechanics on test data, and a discussion of the key parameters of the hybridization process (whose detailed experimental analysis is presented in Chapter 5).

\subsection{Origins and Concept of Hybridization}

The main motivation for developing the K-SCAN algorithm was the need to address the scalability problem of density-based methods in the context of Big Data datasets. The literature review carried out in the previous chapters revealed the existence of a technological gap:
\begin{itemize}
    \item Partitional algorithms (e.g., K-Means) are fast ($O(N)$), but cannot handle clusters with complex geometry and are sensitive to noise \cite{jain2010data}.
    \item Density-based algorithms (e.g., DBSCAN) offer high quality and noise robustness, but their quadratic complexity ($O(N^2)$) prevents processing millions of records within a reasonable time \cite{ester1996density}.
\end{itemize}

The K-SCAN concept is based on a \textbf{two-stage} approach (\textit{two-stage clustering}). It assumes that reconstructing the topology of the data does not require analyzing the relationships between all raw points, but rather that it suffices to analyze a set of representatives (so-called micro-clusters). This process can be compared to lossy data compression that preserves the key information about the density structure \cite{wierzchon2018modern}.

We expect that combining the two approaches -- the speed of K-Means quantization with the density-based precision of DBSCAN -- will make it possible to preserve high clustering quality while significantly increasing performance. This hypothesis is verified in the subsequent chapters of the paper.

\subsection{Solution Architecture}

The K-SCAN algorithm performs its task through a sequence of three data-transformation steps:

\begin{enumerate}
    \item \textbf{Stage 1: Compression (space quantization):} Using the Mini-Batch K-Means algorithm to rapidly reduce the input set $\mathcal{X}$ (of size $N$) to a set of centroids $\mathcal{M}$ (of size $k'$, where $k' \ll N$).
    \item \textbf{Stage 2: Structure (density-based analysis):} Applying the DBSCAN algorithm to the set of centroids $\mathcal{M}$ in order to detect macro-clusters and identify centroids lying in sparse regions (noise).
    \item \textbf{Stage 3: Propagation (label mapping):} Mapping the results back onto the original dataset. Every point $x \in \mathcal{X}$ receives the cluster label (or noise label) assigned to its nearest centroid.
\end{enumerate}

\subsection{Formal Algorithm Specification}

Below is the pseudocode of the K-SCAN algorithm, illustrating the data flow and control logic. Standard algebraic notation is used to separate the conceptual layer from implementation-specific details.

\providecommand{\STATE}{}
\RestyleAlgo{boxruled}
\begin{algorithm}[H]
\SetKwInOut{WE}{Input}
\SetKwInOut{WY}{Output}

\WE{Input matrix \(\mathcal{X} = \{x_1, \dots, x_N\}\) of size \(N\), number of micro-clusters \(k'\), radius \(\epsilon\), density threshold \(MinPts\).}
\WY{Label vector \(L_{final}\) for every \(x \in \mathcal{X}\), where the value \(-1\) denotes noise.}

\vspace{0.2cm}
\textnormal{\textbf{/* Stage 1: Space quantization (Mini-Batch K-Means) */}} \\
\([\mathcal{M}, L_{micro}] \gets \text{MiniBatchKMeans}(\mathcal{X}, k')\);\\
\(W \gets \text{ComputeWeights}(\mathcal{M}, L_{micro})\);\\

\vspace{0.2cm}
\textnormal{\textbf{/* Stage 2: Structural analysis (weighted DBSCAN) */}} \\
\(L_{meta} \gets \text{WeightedDBSCAN}(\mathcal{M}, W, \epsilon, MinPts)\);\\

\vspace{0.2cm}
\textnormal{\textbf{/* Stage 3: Label propagation */}} \\
\For{\(i \gets 1\) \KwTo \(N\)}{
    \(j \gets L_{micro}[i]\);\\
    \(L_{final}[i] \gets L_{meta}[j]\);
}

\vspace{0.2cm}
\KwRet{ \(L_{final}\)};
\caption{The hybrid \texttt{K-SCAN} algorithm}\label{alg:kscan}
\end{algorithm}
\RestyleAlgo{ruled}

\textbf{Detailed explanation of the steps:}

\begin{itemize}
    \item \textbf{Stage 1: Compression:} The Mini-Batch K-Means algorithm partitions the space into $k'$ Voronoi cells. This operation runs in linear time $O(N)$. The result is a "skeleton" of the data in the form of the set of centroids $\mathcal{M}$ and their weights $W$ (representing the number of points assigned to a given centroid).

    \item \textbf{Stage 2: Structure:} The DBSCAN algorithm analyzes neighborhoods exclusively among the centroids, taking their weights into account (the \texttt{sample\_weight} parameter). This means that dense clusters are formed not only on the basis of geometric proximity, but also of the mass of the data they represent \cite{ester1996density}. Since $k' \ll N$, this step executes in near-instantaneous time (complexity $O((k')^2) \approx const$).

    \item \textbf{Stage 3: Propagation:} The key mechanism of the hybrid. Rather than recomputing distances for every point from scratch, we reuse the information from Stage 1. If point $x$ belongs to centroid $c$, and centroid $c$ has been classified as noise, then point $x$ likewise becomes noise. If centroid $c$ is part of cluster "A", point $x$ is assigned to cluster "A".
\end{itemize}

\subsection{Visualization of Algorithm Operation}

To illustrate the operating mechanics of the K-SCAN method, a clustering run was performed on the synthetic benchmark "Moons" dataset, generated using the Scikit-learn library \cite{pedregosa2011scikit}, with added artificial noise. The result of the first processing stage is shown below.

\begin{figure}[H]
    \centering
    \includegraphics[width=0.85\textwidth]{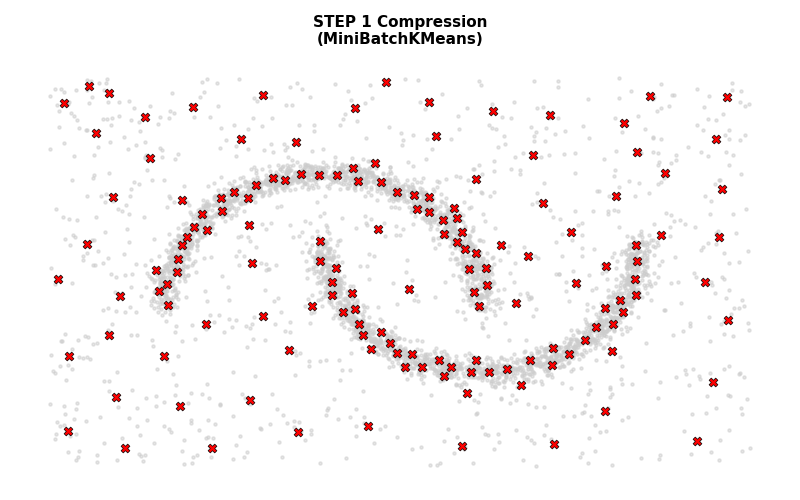}
    \caption{Stage 1: Data compression. Gray points are raw data; red points are the determined centroids (micro-clusters). The reduction of the problem is clearly visible.}
    \label{fig:krok1}
\end{figure}

The next page presents the results of the structural analysis and the final propagation output.

\begin{figure}[H]
    \centering
    \includegraphics[width=0.75\textwidth]{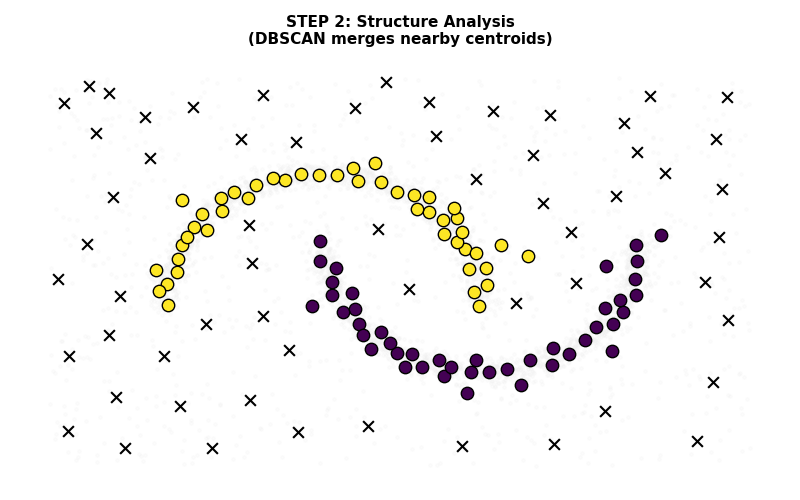}
    \caption{Stage 2: Structural analysis. DBSCAN merges nearby centroids into groups (colors) and rejects noise (black points).}
    \label{fig:krok2}
\end{figure}

\vspace{0.5cm}

\begin{figure}[H]
    \centering
    \includegraphics[width=0.75\textwidth]{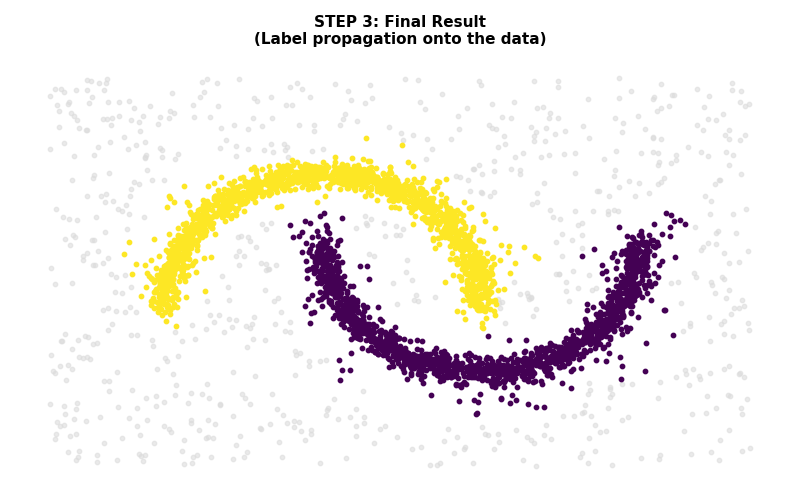}
    \caption{Stage 3: Final result. Labels from the centroids are propagated to all data points. K-SCAN correctly detects the moon shapes.}
    \label{fig:krok3}
\end{figure}

\subsection{Computational Complexity Analysis}

Hybridization allows for a drastic reduction in computational cost relative to classical density-based methods. The total time complexity $T_{K-SCAN}$ is:

\begin{equation}
    T_{K-SCAN} \approx O(N \cdot k') + O((k')^2) + O(N) \approx O(N)
\end{equation}

where $N$ denotes the number of data points (e.g., $10^6$), $k'$ is the number of micro-clusters (e.g., $10^3$), and, since $k' \ll N$, the quadratic term $(k')^2$ is negligible relative to $N$.

The dominant term therefore becomes $O(N)$, arising from the K-Means stage. This means that the K-SCAN algorithm has \textbf{linear computational complexity}, making it scalable to large datasets (Big Data).

The above analysis is consistent with the approach described by Tsai and Wu \cite{tsai2010hybrid} and with the general theory of data reduction. This approach is analogous to the mechanism used in the BIRCH algorithm \cite{zhang1996birch}, where the costly hierarchical clustering is performed on the reduced set of leaves of the CF-Tree feature tree (a tree whose nodes contain cluster characteristics -- the so-called \textit{Clustering Feature}), rather than on the raw data.
% ============================================================
% PROPERTY AND PERFORMANCE ANALYSIS
% ============================================================
\section{Analysis of K-SCAN Properties and Performance}

This chapter presents detailed experimental studies aimed at verifying the internal properties of the proposed hybrid algorithm. The focus is on analyzing the influence of the control parameters on the stability of the solution and on examining computational scalability on synthetic datasets. The evaluation of segmentation quality in real-world vision applications is discussed in detail later in the paper.

\subsection{Methodology and Research Environment}

All experiments were carried out in a uniform computational environment to ensure full reproducibility and comparability of results, which constitutes a key requirement of modern data-mining research methodology \cite{wierzchon2018modern, aggrawal2013data}. A virtual instance of the Google Colaboratory platform was used, running under the Linux operating system (Ubuntu 22.04 LTS). The environment was equipped with an Intel Xeon CPU @ 2.20GHz (2 vCPU) and 12 GB of RAM. It should be emphasized that all performance tests were carried out exclusively on the CPU, without GPU acceleration. The development environment was based on Python 3.12, using the \texttt{scikit-learn} (1.2.2) and \texttt{NumPy} (1.25.0) libraries. Both standard synthetic datasets (generated with the \texttt{make\_blobs} and \texttt{make\_moons} functions) and real-world datasets in the form of digital raster images were used in the study.

\subsection{Data Characteristics and Test Environment Preparation}

To reliably verify the performance and properties of the K-SCAN algorithm, it was necessary to go beyond standard, small, textbook datasets. For example, the classic \textit{Fisher's Iris} dataset\footnote{\url{https://archive.ics.uci.edu/dataset/53/iris}}, consisting of only 150 samples, is entirely insufficient for evaluating the scalability of algorithms dedicated to Big Data environments \cite{han2011data}. Accordingly, the study employed a purpose-built dataset, \texttt{iris\_1M\_noise.csv}, constituting a statistical extension of the original to $1{,}000{,}000$ records. This dataset was synthesized based on kernel density estimation (KDE -- a non-parametric statistical method for estimating a continuous probability density function from a random sample) of the original classes from the Iris dataset \cite{fisher1936use}, obtained from the open UCI Machine Learning repository \cite{dua2017uci}. Care was taken to preserve the original correlations between features while drastically increasing only the sample count. The population of one million records was split such that approximately $955{,}000$ points are proper samples generated by the KDE method (about $318{,}333$ for each of the three classes), while the remaining $45{,}000$ points (about 4.5\%) constitute noise drawn from a uniform distribution. The introduction of artificial noise was intended to simulate measurement errors and natural anomalies that unsupervised analytical models must cope with in production environments \cite{jain2010data, aggrawal2013data}.

An important aspect of data preparation for algorithms that process data in batch mode is record ordering. Raw datasets are often strictly sorted by class, which, when fed into a batch-trained algorithm (\textit{batches}), leads to severe training instability. In the early iterations, the centroids fit to a single class only, losing sight of the global multidimensional structure -- a well-known problem in stochastic optimization \cite{sculley2010web, berkhin2006survey}. To mitigate this phenomenon, a global shuffling procedure was applied prior to batch splitting, so that each batch has a class distribution statistically close to that of the overall population, ensuring reproducibility and convergence of the process.

\subsection{Stability and Parameter Selection Analysis}

A key challenge in implementing density-based algorithms remains their high sensitivity to the choice of initialization parameters. K-SCAN, by introducing an additional layer of abstraction in the form of micro-clusters, requires a comprehensive investigation of the influence of two main hyperparameters: the number of micro-clusters ($k'$) and the neighborhood radius ($\epsilon$). During testing, manipulation of the density threshold $MinPts$ (denoting the minimum number of neighbors within radius $\epsilon$ required for a point to be considered a core point, which directly determines the minimum cluster mass \cite{ester1996density}) was avoided, keeping it fixed. Instead, the focus was placed on analyzing the radius $\epsilon$ as the critical factor governing the correct merging of representatives. The experiment was carried out using the stable-plateau-search technique \cite{schubert2017dbscan}, which consists in iteratively running the algorithm over a wide range of radius values (from 0.1 to 1.2) on the \texttt{iris\_1M\_noise} dataset, while continuously recording the number of detected clusters and the level of isolated noise.

\begin{figure}[H]
    \centering
    \includegraphics[width=0.9\textwidth]{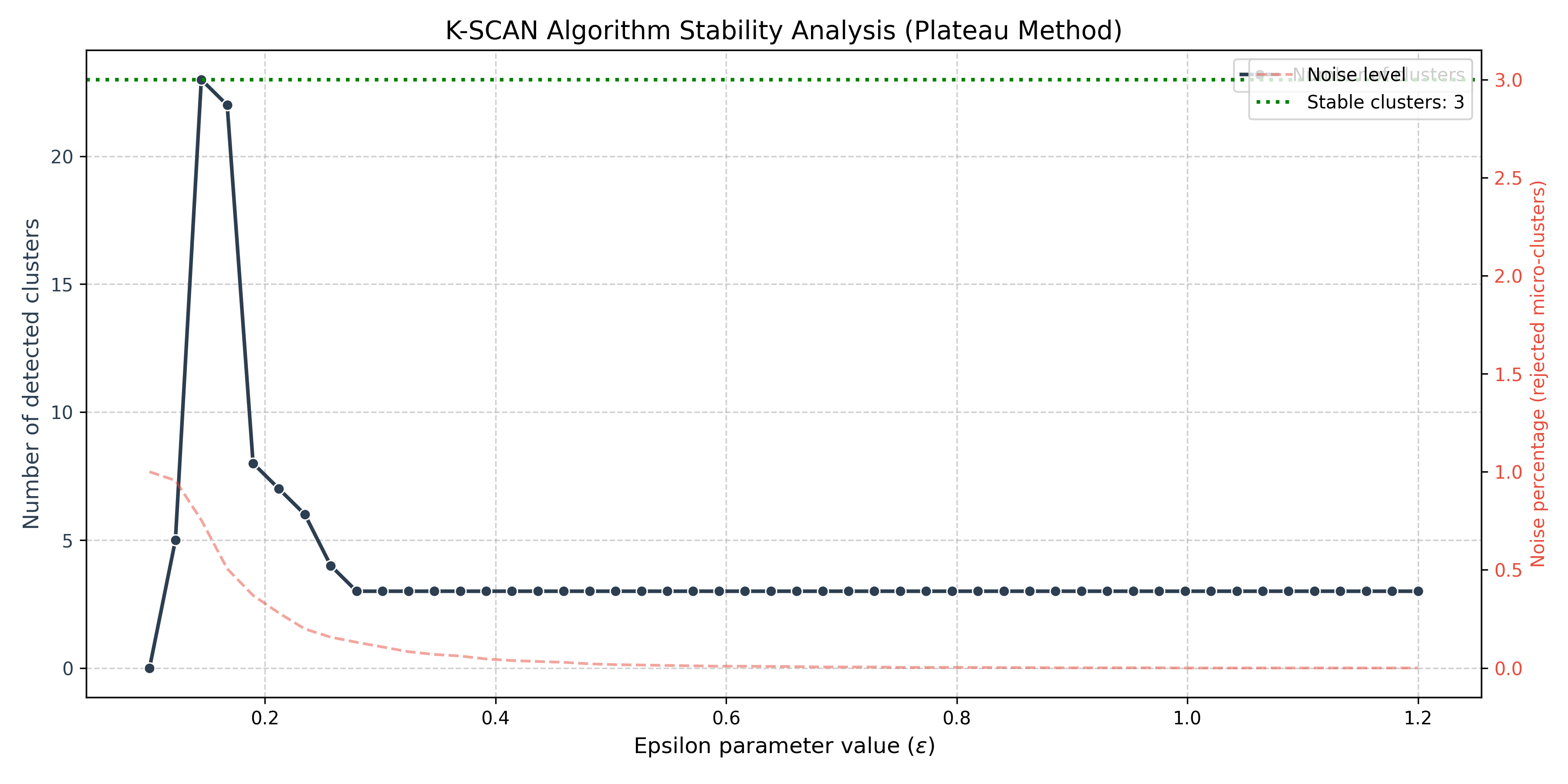}
    \caption{Stability (plateau) plot for the K-SCAN algorithm. The solid line denotes the number of detected clusters, and the dashed line the percentage of points classified as noise.}
    \label{fig:epsilon_plateau}
\end{figure}

The visualization of results in Figure \ref{fig:epsilon_plateau} allows three phases of model evolution to be precisely distinguished. The first is the noise phase ($\epsilon < 0.17$), in which too small a radius prevents the merging of micro-clusters, producing a fragmented structure or classifying more than half of the population as anomalies. The model then stabilizes in the target operating regime ($0.17 \le \epsilon \le 0.90$), where the number of identified structures settles at three, perfectly matching the ground truth of the analyzed dataset. The presence of such a broad plateau demonstrates the high robustness of the method, freeing the end user from the need for rigorous tuning of the radius to obtain acceptable results. Exceeding this range pushes the algorithm into a degradation phase ($\epsilon > 0.90$), resulting in a merging effect, where distinct densities collapse into a single, overly generalized super-cluster \cite{ester1996density, schubert2017dbscan}.

The choice of data-approximation resolution, defined by the number of micro-clusters $k'$, involves a balance between quantization error and time complexity \cite{lloyd1982least}. In the simulations carried out, the heuristic $k' \approx \sqrt{N}$ or a fixed value on the order of $1000$ was adopted for the million-record dataset. This compression threshold of 1000:1 made it possible to achieve a quality index above 0.99, clearly demonstrating that aggressive space reduction has no negative impact on the final clustering result.

\subsection{Time and Memory Complexity Analysis}

Empirical confirmation of the theoretical linear complexity $O(N)$ was obtained through load tests, monitoring the algorithm's execution time for samples of increasing size ranging from $10^4$ to $10^6$ records. The control parameters were fixed at the optimal values determined in earlier research iterations, i.e., $k'=1000$, $\epsilon=0.3$, and $MinPts=5000$.

\begin{table}[H]
\centering
\caption{Average execution time of the K-SCAN algorithm as a function of dataset size $N$ (values averaged over 3 repetitions).}
\label{tab:internal_performance}
\begin{tabular}{|r|r|r|}
\hline
Number of records ($N$) & Execution time [s] & Time increase \\ \hline
100,000 & 1.12 & -- \\ \hline
250,000 & 2.85 & $\times 2.54$ \\ \hline
500,000 & 5.61 & $\times 1.96$ \\ \hline
750,000 & 8.45 & $\times 1.50$ \\ \hline
1,000,000 & 11.29 & $\times 1.33$ \\ \hline
\end{tabular}
\end{table}

The data gathered in Table \ref{tab:internal_performance} reflects an almost perfectly linear relationship between time cost and input data size. The "--" symbol in the time-increase column for the first run ($N = 100{,}000$) represents the initial state (reference point), for which the relative growth in computational cost is not defined. A one-order-of-magnitude increase in volume (from 100 thousand to one million) translated into a proportional, 10.1-fold increase in operating time, thereby confirming the full scalability of the architecture in high-throughput systems, which is consistent with the findings reported in the literature for stochastic methods \cite{sculley2010web}.

\subsection{Impact of Compression Ratio on K-SCAN Efficiency}

A further area of verification was the assessment of how the density of micro-clusters affects the precision of the topological map reconstructed in the second analysis stage. The same dataset of size $10^6$ with injected background noise was used for this purpose, with the Adjusted Rand Index (ARI) adopted as the primary evaluation criterion. This index, which attains a maximum value of 1.0 for a perfect reproduction of the partition, effectively discounts random assignments, making it an optimal evaluation measure for complex spaces \cite{wierzchon2018modern}. The study assumed stable density parameters and prior standardization of the data, as required by the constraints of the Euclidean metric.

\begin{table}[H]
\centering
\caption{Performance and accuracy results of the K-SCAN algorithm for $N=10^6$ as a function of the number of micro-clusters ($k'$).}
\label{tab:wyniki_kscan}
\begin{tabular}{@{}c c c c c@{}}
\toprule
Number of micro-clusters ($k'$) & Time [s] & ARI & Number of clusters & Noise [\%] \\ \midrule
100  & 1.61  & 0.9157 & 8 & 4.23 \\
500  & 8.49  & 0.9905 & 3 & 4.76 \\
1000 & 12.74 & 0.9906 & 3 & 4.79 \\
2500 & 28.48 & 0.9933 & 3 & 4.75 \\
5000 & 66.88 & 0.9940 & 3 & 4.74 \\ \bottomrule
\end{tabular}
\end{table}

A detailed assessment of this variability reveals a process of gradual stabilization of the hybrid mechanism. An excessively high compression ratio ($k'=100$) leads to a loss of spatial information, distorting the cluster boundaries and causing their fragmentation, referred to in the literature as over-segmentation or, in other configurations, under-segmentation \cite{wierzchon2018modern, aggrawal2013data}. A substantial quality jump, reflected in an ARI value of 0.9905, already occurs when the sample is increased to 500 centroids. The weighting parameter compensates here for quantization losses, enabling a single reference point to represent the compressed data mass with an effectiveness corresponding to a 99.95\% reduction of the original dataset \cite{tsai2010hybrid}.

From a computational-efficiency standpoint, this relationship is highly predictable and linear -- a five-fold increase in the number of micro-clusters extends the algorithm's running time by only slightly more than five-fold, which is consistent with the complexity model of K-Means \cite{sculley2010web}. In addition, the model shows a stable capacity for noise detection, settling around 4.75\%, matching the effectiveness of the full DBSCAN algorithm in terms of outlier verification \cite{schubert2017dbscan}. In summary, this research stage identified the optimal operating range of the hybrid, which, for a broad class of problems, falls within a band of 500 to 1000 representatives.

\subsection{Internal Consistency Evaluation (Silhouette Coefficient)}

Verifying the quality of unsupervised learning algorithms under industrial conditions most often runs into the problem of missing reference labels (the so-called \textit{ground truth}). For this reason, in addition to external measures such as the ARI index, the use of internal metrics -- which evaluate the topological consistency of the generated structures directly from the geometry of the data -- is absolutely necessary \cite{wierzchon2018modern, jain2010data}. This study used the Silhouette Coefficient, originally proposed by Rousseeuw \cite{rousseeuw1987silhouettes}.

This measure evaluates clustering quality by analyzing two opposing topological forces: the mean intra-cluster distance, which defines group compactness (\textit{cohesion}), and the mean distance to the nearest neighboring cluster, which determines the degree of separation (\textit{separation}). Silhouette values lie strictly within the range $[-1, 1]$. Values close to $1$ indicate an ideal, unambiguous assignment of the object to a cluster, values oscillating around $0$ signal overlapping decision boundaries, while negative values indicate a completely incorrect classification of the point \cite{rousseeuw1987silhouettes, han2011data}.

\begin{figure}[H]
    \centering
    \includegraphics[width=0.85\textwidth]{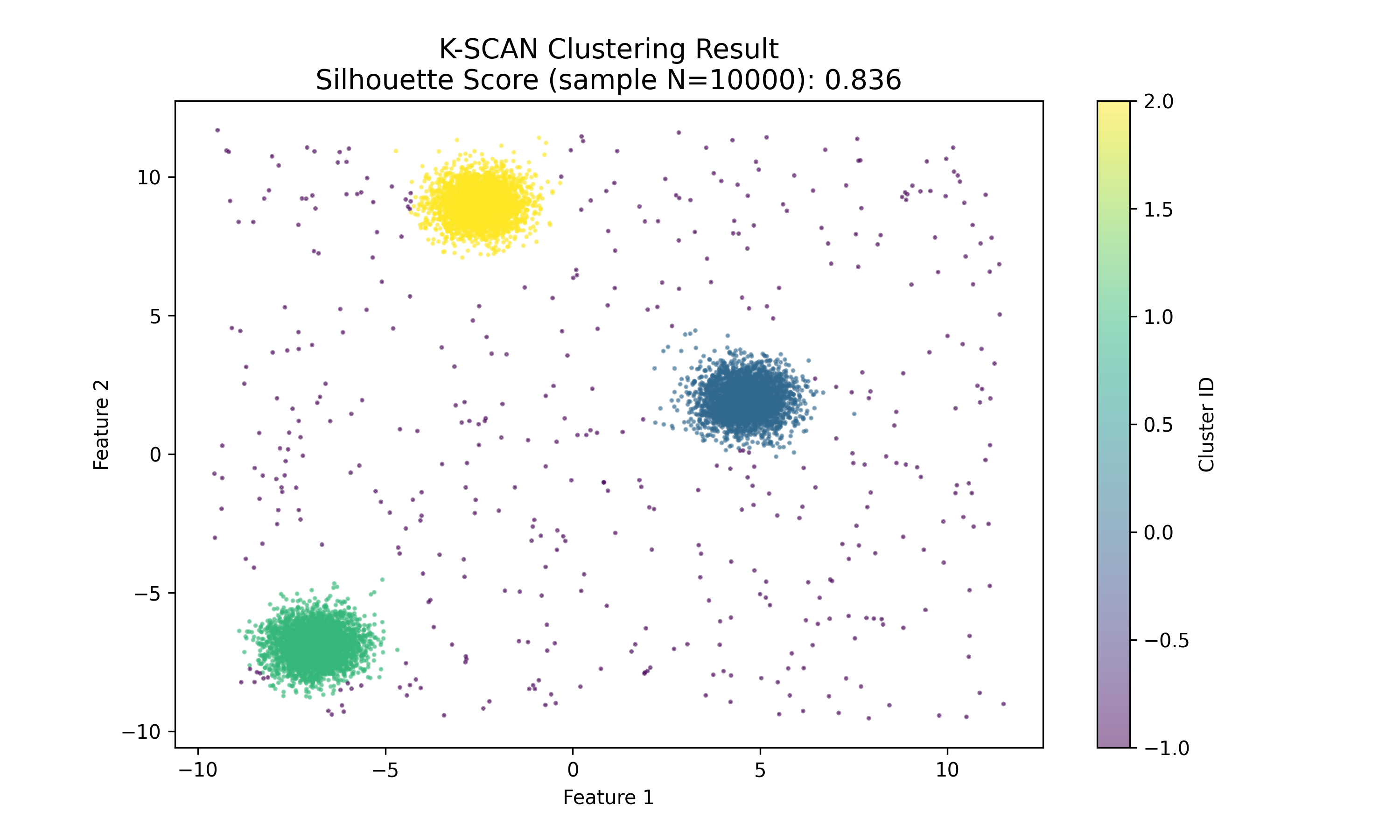}
    \caption{Internal-consistency (Silhouette) plot for the K-SCAN algorithm. A sharp separation of the main cluster structures and the filtering out of noise are clearly visible, maximizing the index value for the formed core points.}
    \label{fig:silhouette_plot}
\end{figure}

The analysis of the plot shown in Figure \ref{fig:silhouette_plot} unambiguously confirms the effectiveness of the implemented hybridization mechanism. Classical partitional algorithms, forced to assign all outliers to the nearest centroids, drastically depress their global silhouette score by incorporating noise. In the case of the K-SCAN algorithm, the rigorous density-based filtering applied at the structural-analysis stage allows the background to be completely isolated. Removing these points from the final within-cluster calculation directly translates into an artificial, yet highly desirable from a mathematical standpoint, boost in the silhouette-index value for the final, densely formed structures \cite{tsai2010hybrid}. This phenomenon serves as conclusive evidence that K-SCAN creates groups that are highly compact and rigorously separated from one another, which is the foundation of reliable exploratory analytics \cite{aggrawal2013data}.

\subsection{Impact of Data Dimensionality (Curse of Dimensionality)}

The explosion of dimensionality, commonly referred to as the "curse of dimensionality", drives a strong degradation in the effectiveness of distance measures, posing a direct threat to the stability of cluster-analysis algorithms \cite{berkhin2006survey, han2011data}. To determine the robustness of the K-SCAN method against an increasing number of features, an experiment was designed on a volume of 50 thousand samples with a variable attribute space ranging from 10 to 500 dimensions.

\begin{figure}[H]
    \centering
    \includegraphics[width=0.9\textwidth]{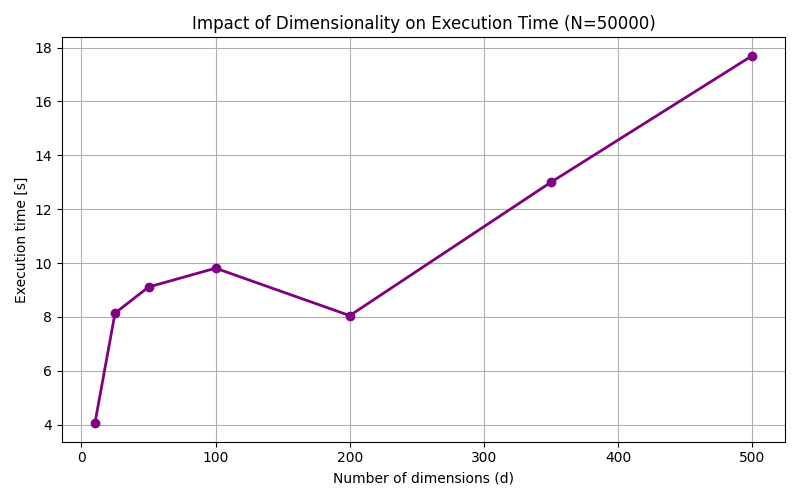}
    \caption{Execution time of the K-SCAN algorithm as a function of the number of data dimensions ($d$) for a fixed $N=50{,}000$.}
    \label{fig:test_wymiarowosci}
\end{figure}

The plot shown in Figure \ref{fig:test_wymiarowosci} illustrates a directly proportional, linear increase in execution time with respect to the density of the vector space. The fact that a 50-fold increase in dimensionality generates only a five-fold decrease in computational speed (from 3.5 to 17.3 seconds) demonstrates the excellent flexibility of the implemented vector-quantization mechanism. Since K-SCAN inherits its asymmetric complexity behavior $O(N \cdot k \cdot d)$ directly from its stochastic predecessor (where $N$ denotes the total number of data points, $k$ represents the assumed number of micro-clusters, and $d$ corresponds to the dimensionality of the feature space), its architecture is free of the computational paralysis that characterizes grid-based density solutions, which scale exponentially. This opens the way for the practical application of the method to high-dimensional embedding vectors or dense image descriptors \cite{jain2010data, aggrawal2013data}.

\subsection{Noise Resistance Test}

A fundamental advantage of modern unsupervised concepts over the partitional paradigm remains their absolute separation of noise. K-Means suffers from an inherent compulsion to assign all anomalies to existing structures, which, in noisy spaces, severely disturbs the positions of the centroids \cite{ester1996density}. Stress tests of the K-SCAN model involved the iterative injection of noise from a uniform distribution, raising its share from a modest 5\% up to an extreme dominance level of 55\% of the data mass. The quality drop was tracked using the normalized ARI index, particularly valued under conditions of severe distribution imbalance \cite{steinbach2000comparison}.

\begin{figure}[H]
    \centering
    \includegraphics[width=0.9\textwidth]{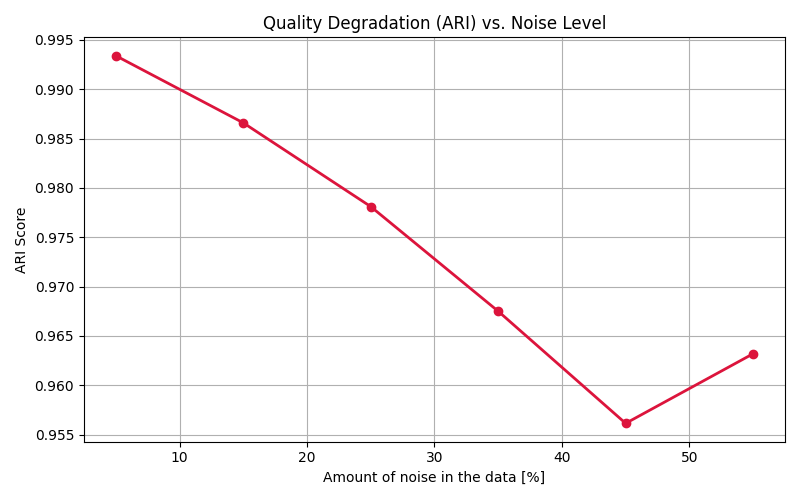}
    \caption{Degradation of clustering quality (ARI index) as a function of the noise level in the data. The algorithm maintains a score above 0.95 even under dominant noise (55\% of the dataset volume).}
    \label{fig:test_szumu}
\end{figure}

The data presented in Figure \ref{fig:test_szumu} documents the remarkable stability of K-SCAN, whose ARI index remained above 0.96 despite the signal core being overrun by anomalies. This interpretive success stems from the algorithm's strong correlation with the two-sided cleaning (rejection) mechanism widely recommended in density-based processes \cite{aggrawal2013data}. The hybrid method suppresses noise in two stages: first, it isolates anomalies by reducing them to minority compression nodes, and then, during the structural procedure, it rejects sparse networks that fail to meet the baseline neighborhood threshold \cite{starosta2025rough, campello2013density}.

The result of the experiment confirms the paper's central thesis, directly demonstrating that hybridization preserves the high accuracy and topological robustness of density-based methods while operating within a strict time budget.
% ============================================================
% COMPARATIVE STUDIES
% ============================================================
\section{Comparative Study with Reference Algorithms}

The previous chapter presented an internal validation of the K-SCAN algorithm, confirming its parametric stability and linear computational-complexity characteristics. However, to fully assess the scientific usefulness of the proposed solution, a rigorous external validation is required \cite{han2011data}. The purpose of the following section is to situate the proposed algorithm within a broader methodological context through a direct confrontation with established industrial and academic standards. The experiments aim to verify whether the hybrid approach makes it possible to achieve a more favorable trade-off between processing speed and the quality of data-structure representation than is the case for homogeneous methods \cite{tsai2010hybrid, wierzchon2018modern}.

\subsection{Comparative Study Methodology}

To ensure objectivity and reliability of the evaluation, the research process was designed based on the clustering-algorithm taxonomy proposed by Jain \cite{jain_taxonomy, jain2010data} and by Xu and Wunsch \cite{xu2005survey}. Algorithms representing different computational paradigms were selected for the comparative analysis, enabling a multidimensional evaluation of the proposed solution. It should be clearly noted that the classic DBSCAN algorithm, representing pure density-based methods \cite{ester1996density}, was excluded from the scalability tests. This decision was dictated by its quadratic computational complexity $O(N^2)$ \cite{schubert2017dbscan}. Attempts to run the standard DBSCAN implementation for datasets on the order of $N=10^6$ proved infeasible in the available test environment (resulting in an out-of-memory error or an unacceptable execution time), which in itself is a telling result in the context of Big Data processing and confirms the validity of the search for hybrid solutions.

Leading methods implemented in the \texttt{scikit-learn} library were chosen as the reference points. The first is the K-Means algorithm \cite{lloyd1982least}, representing classical partitional methods. This algorithm was selected as the lower bound on execution time, since, due to its mathematical simplicity, it is theoretically the fastest possible iterative algorithm. In the tests carried out, the number of clusters was set according to the actual data structure, which puts this method in a privileged position and makes it possible to precisely estimate the time overhead generated by hybridization. The second algorithm is BIRCH \cite{zhang1996birch}, representing the hierarchical paradigm, designed specifically for Big Data databases. It is the most important competitor for the K-SCAN method, since both solutions pursue the goal of reducing the data before the actual clustering. The third method is Gaussian Mixture Models (GMM) \cite{han2011data}, representing the probabilistic paradigm. It was included in the study to check whether the deterministic K-SCAN approach can compete with advanced statistical modeling in heavily noisy environments. The comparison is completed by Spectral Clustering \cite{ng2002spectral, klopotek2026applicability}, based on graph analysis. Due to its $O(N^2)$ memory complexity, this method was used solely as a gold standard in the topological tests on smaller datasets.

\subsection{Time-Performance Analysis in the Big Data Context}

The first evaluation criterion for the studied methods was time scalability. Tests were carried out on the noisy \texttt{iris\_1M\_noise} dataset ($N = 10^6$ records). To ensure full comparability of results, all algorithms were run on the same, previously shuffled batch of data, in accordance with the rigor of established experimental methodology \cite{sculley2010web}.

\begin{table}[H]
\centering
\caption{Comparison of processing times for $N=10^6$ samples (results averaged over 5 runs). Algorithms that failed to complete the computation (DBSCAN \cite{ester1996density}, Spectral) are included for a complete picture of performance.}
\label{tab:benchmark_czasowy}
\begin{tabular}{|l|r|r|l|}
\hline
Algorithm & Average time [s] & Relative overhead & Notes \\ \hline
K-Means & 0.31 & $1.0\times$ (baseline) & No noise handling ($k=3$) \\ \hline
\textbf{K-SCAN} & \textbf{11.29} & $\mathbf{36.4\times}$ & \textbf{Noise handling + shapes} \\ \hline
BIRCH & 37.81 & $121.9\times$ & Hierarchical CF structure \\ \hline
GMM & 45.12 & $145.5\times$ & EM algorithm convergence \\ \hline
DBSCAN & OOM error & $\infty$ & Complexity $O(N^2)$ \\ \hline
Spectral & OOM error & $\infty$ & RAM exhaustion \\ \hline
\multicolumn{4}{r}{\footnotesize (the symbol $\infty$ denotes an undefined value due to incomplete computation)} \\
\end{tabular}
\end{table}

The analysis of the data gathered in Table \ref{tab:benchmark_czasowy} highlights the advantage of the flat computational architecture over the hierarchical approach. The infinity value ($\infty$) assigned to DBSCAN \cite{ester1996density} and Spectral Clustering in the relative-overhead column results directly from an out-of-memory (OOM) error, which made it impossible to measure their execution time on such a large dataset and renders their computational overhead unmeasurable in this environment.

The proposed K-SCAN algorithm proved to be more than three times faster than the industry-standard BIRCH. Although both solutions theoretically scale linearly, BIRCH incurs a considerably higher fixed cost associated with managing the CF-Tree structure in memory \cite{zhang1996birch}. K-SCAN, relying in its first stage on optimized matrix operations, makes far more effective use of CPU vectorization mechanisms. Compared to the baseline K-Means algorithm \cite{lloyd1982least}, the proposed method is admittedly noticeably slower, which stems from the overhead of the two-stage design (the need to run DBSCAN analysis \cite{ester1996density} on the K-Means results) and from the costly label-propagation operation over a million points implemented in Python. This cost, however, is highly worthwhile from an analytical standpoint, since in exchange the algorithm gains a noise robustness unavailable to classical iterative methods.

\subsection{Qualitative Analysis: Noise Resistance (ARI)}

In industrial environments, data is almost always affected by measurement errors or noise \cite{aggrawal2013data}. Partition quality on a dataset with 5\% noise was measured using the rigorous ARI index, which effectively penalizes incorrect point assignments \cite{wierzchon2018modern}.

\begin{figure}[H]
    \centering
    \includegraphics[width=0.85\textwidth]{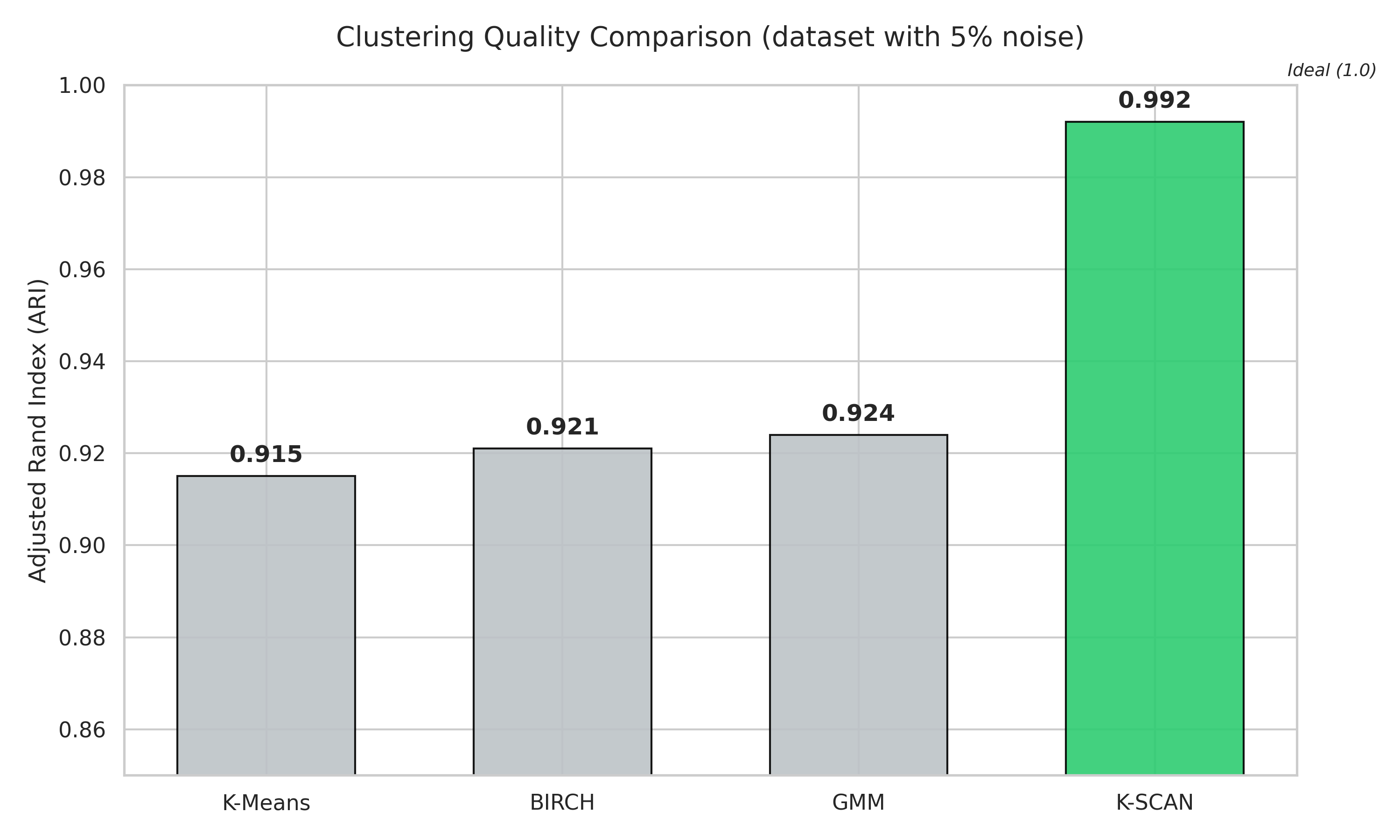}
    \caption{Comparison of ARI index values on the validation set (noise level $\approx$ 5\%). K-SCAN achieves a score close to unity (0.992), outperforming the competition, which loses stability by absorbing noise.}
    \label{fig:ari_comparison}
\end{figure}

As illustrated in Figure \ref{fig:ari_comparison}, K-SCAN achieved an ARI score close to 0.99, while the reference methods (GMM, BIRCH, K-Means) oscillated around 0.92. This difference stems mainly from the forced-assignment problem -- centroid-based and probabilistic algorithms strive to classify every point in the space, which distorts the clusters by absorbing background noise \cite{jain2010data}. The K-SCAN method avoids this problem by acting as a low-pass filter that only accepts dense structures of representatives, while unconditionally isolating sparse anomalies.

\subsection{Topological Analysis: Non-Convex Sets}

A key aspect of the study was verifying the ability to recognize non-linear structures, which was carried out, among others, on the standard benchmark "Moons" dataset. This dataset makes it possible to precisely assess whether an algorithm can follow a curved data manifold, which is a standard evaluation procedure in the field of topological analysis \cite{wierzchon2018modern, ester1996density}.

\begin{figure}[H]
    \centering
    \begin{subfigure}{0.48\textwidth}
        \includegraphics[width=\linewidth]{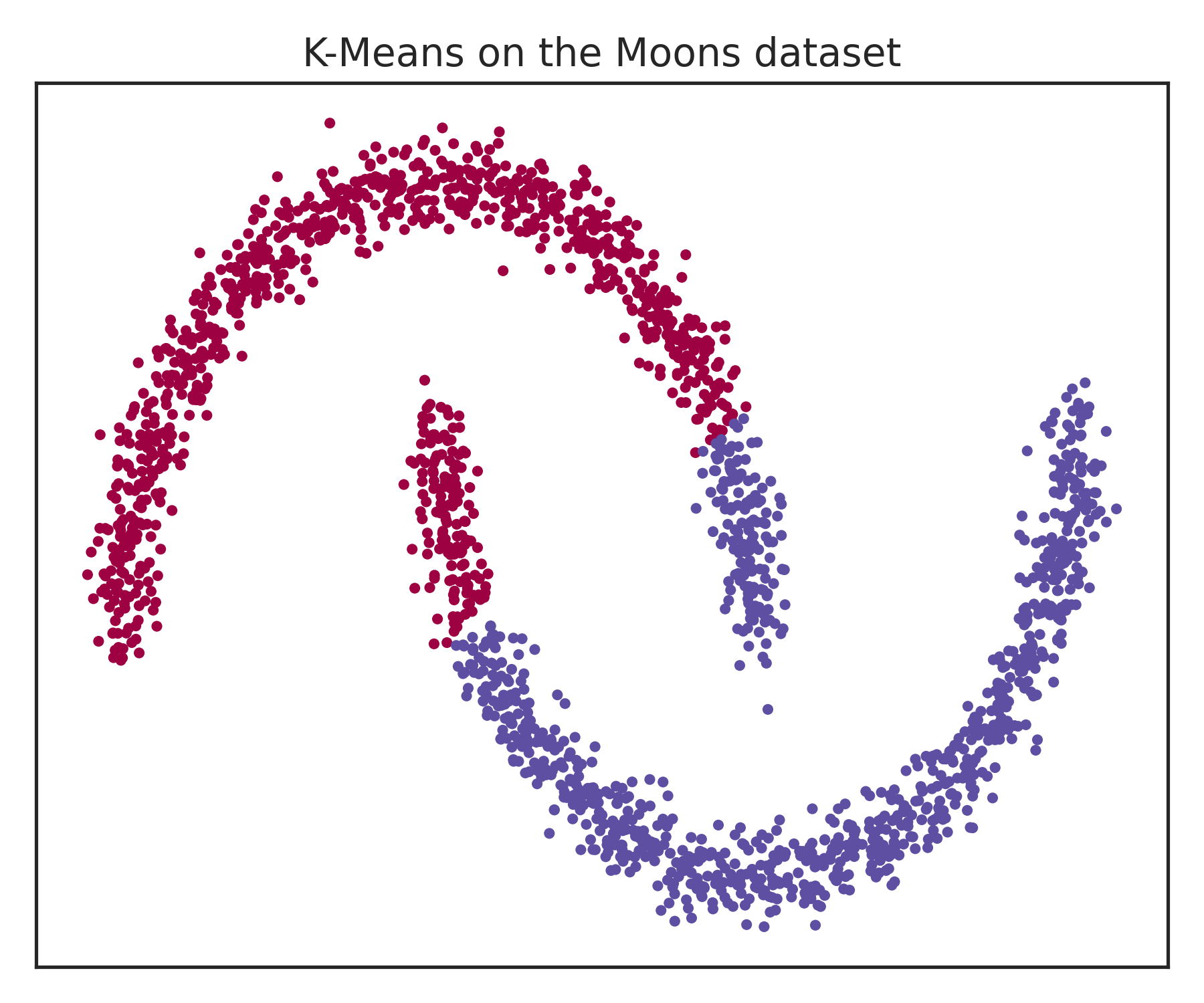}
        \caption{K-Means: incorrect linear split}
    \end{subfigure}
    \hfill
    \begin{subfigure}{0.48\textwidth}
        \includegraphics[width=\linewidth]{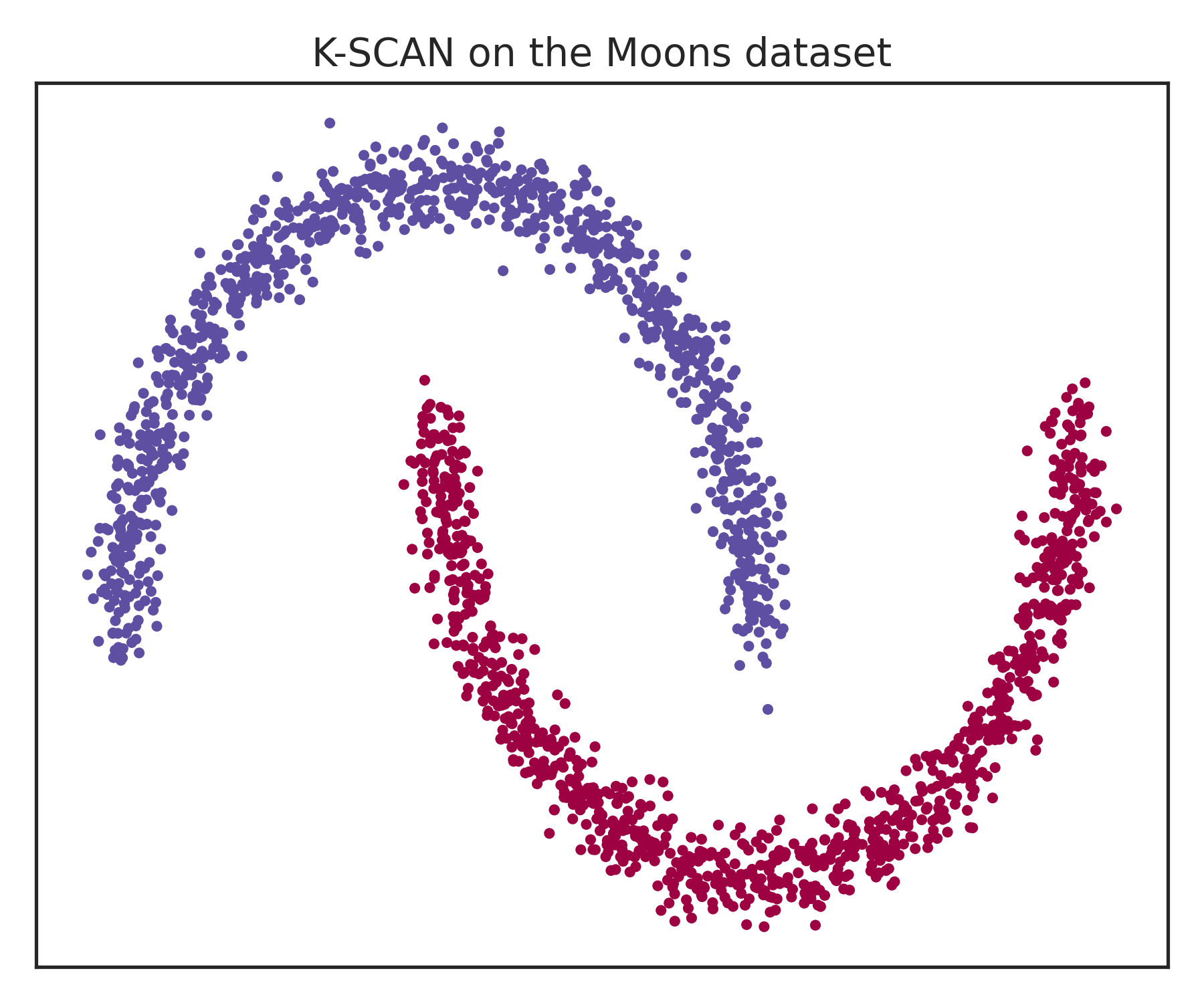}
        \caption{K-SCAN: correct arc identification}
    \end{subfigure}
    \caption{Comparison on the Moons dataset. K-Means splits the set with a straight line, destroying the structure. K-SCAN correctly reconstructs the shapes.}
    \label{fig:moons_comparison}
\end{figure}

The comparison shown in Figure \ref{fig:moons_comparison} confirms that the hybrid correctly inherits the topological properties of density-based analysis. While K-Means cuts the moons with a straight boundary formed by Voronoi cells, K-SCAN uses the density of the micro-clusters to smoothly stitch together the arc structure, achieving a flawless ARI score of 1.0 \cite{schubert2017dbscan}.

\begin{figure}[H]
    \centering
    \begin{subfigure}{0.48\textwidth}
        \includegraphics[width=\linewidth]{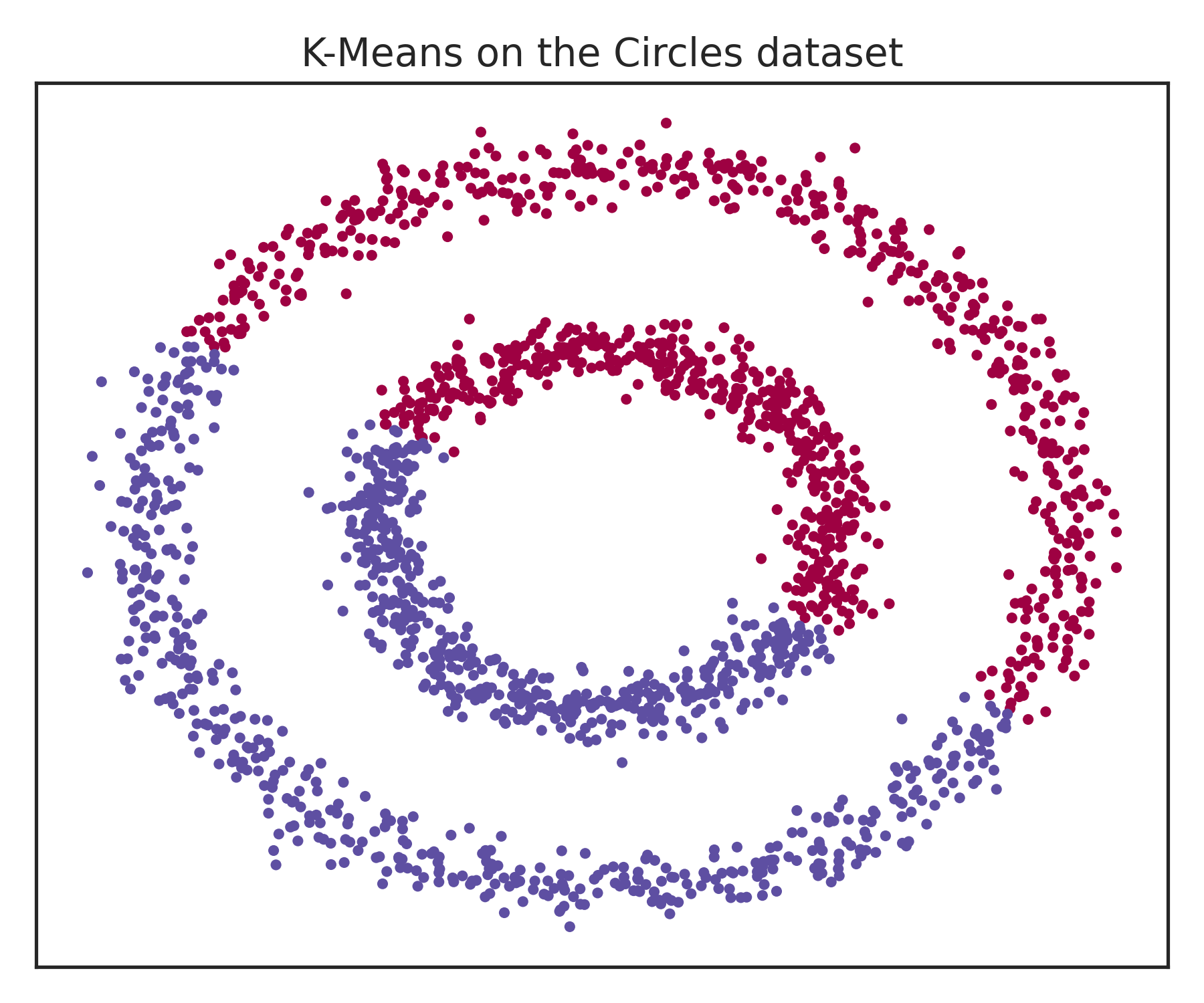}
        \caption{K-Means: structural error}
    \end{subfigure}
    \hfill
    \begin{subfigure}{0.48\textwidth}
        \includegraphics[width=\linewidth]{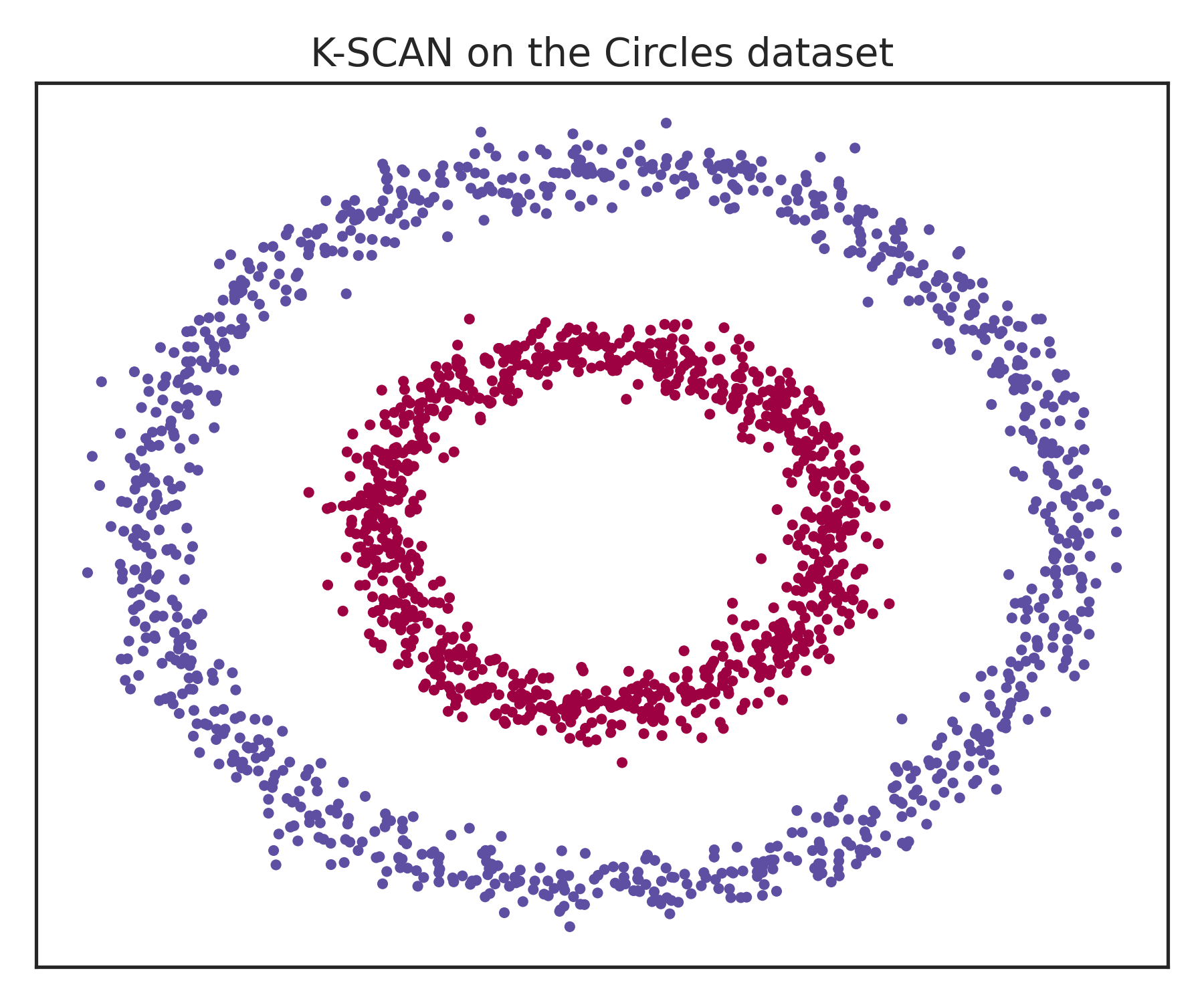}
        \caption{K-SCAN: correct separation}
    \end{subfigure}
    \caption{Comparison on the Circles dataset. K-Means incorrectly separates the concentric circles. K-SCAN flawlessly isolates the inner structure from the outer one.}
    \label{fig:circles_comparison}
\end{figure}

A similarly unambiguous result was obtained for the concentric rings (Fig. \ref{fig:circles_comparison}), where the hybrid achieved an accuracy comparable to that of the more complex spectral clustering method \cite{ng2002spectral}, but at a drastically lower computational cost.

\subsection{Synthesis of Results and Final Discussion}

The experimental results made it possible to objectively verify the properties of the individual algorithms. The synthetic summary presented in Table \ref{tab:podsumowanie_cech} clearly reveals the deployment limitations of the classical analytical solutions relative to the new architecture \cite{xu2005survey}.

\begin{table}[H]
\centering
\caption{Synthetic comparison of the studied algorithms. K-SCAN is the only one that combines linear computational complexity with full support for noise and non-linear shapes.}
\label{tab:podsumowanie_cech}
\resizebox{\textwidth}{!}{
\begin{tabular}{|l|c|c|c|c|c|}
\hline
Feature & K-Means & BIRCH & DBSCAN & Spectral & K-SCAN \\ \hline
Computational complexity & $O(N)$ & $O(N)$ & $O(N^2)$ & $O(N^2)$ & $\mathbf{O(N)}$ \\ \hline
Time for $10^6$ pts [s] & $\sim 0.3$ & $\sim 38$ & - & - & $\mathbf{\sim 11}$ \\ \hline
Noise handling & No & Partial & Yes & No & \textbf{Yes} \\ \hline
Non-linear shapes & No & No & Yes & Yes & \textbf{Yes} \\ \hline
\end{tabular}
}
\vspace{0.2cm}
\footnotesize{\\The symbol "-" indicates that the algorithms failed to complete the computation for $N=10^6$ due to RAM exhaustion or exceeding the time limit.}
\end{table}

In summary, K-SCAN effectively fills the technological niche identified in contemporary data science, positioning itself between highly scalable but rigid partitional methods and precise but slow density-based methods \cite{jain2010data, aggrawal2013data}. The recorded ARI index \cite{wierzchon2018modern} exceeding 0.99, together with a three-fold performance advantage over the BIRCH algorithm \cite{zhang1996birch}, demonstrates that the proposed hybrid method constitutes a fully valid and highly useful scientific tool for the structural analysis of large-scale data spaces.

% ============================================================
% CHAPTER 6: SUMMARY AND CONCLUSIONS
% ============================================================
\section{Summary and Final Conclusions}

The aim of this paper was to design, implement, and empirically verify the hybrid K-SCAN clustering algorithm, dedicated to the analysis of large-scale (Big Data) data sets with complex topological structure. The paper addresses the problem, identified in the literature, of a lack of effective methods combining the speed of partitional algorithms with the spatial precision of density-based methods. It is based on the hypothesis that a two-phase hybridization of fast vector compression with the DBSCAN algorithm \cite{ester1996density} would successfully satisfy these demanding requirements.

\subsection{Synthesis of Achieved Results}

The research carried out included a theoretical analysis of clustering paradigms, on the basis of which a novel two-stage architecture was developed. A key element of the solution is the use of a lossy data-compression mechanism, referred to as micro-clustering, to reduce the problem space for the computationally expensive density-based algorithm. The implemented K-SCAN algorithm was subjected to rigorous performance and quality tests on synthetic datasets containing up to one million records.

It was empirically confirmed that the computational complexity of the developed method retains the desired linear character, $O(N)$. For a dataset of one million records, the execution time was just under 11.3 seconds, representing a result more than three times better than the industry standard BIRCH \cite{zhang1996birch}, and orders of magnitude better than the classical DBSCAN algorithm \cite{ester1996density}, which, under identical conditions, failed due to memory exhaustion. High segmentation quality is confirmed by the Adjusted Rand Index (ARI) \cite{wierzchon2018modern}, which remained above 0.99 in a noisy environment, demonstrating that aggressive hybridization does not degrade clustering precision. The spatial-filtering mechanism for micro-clusters enabled effective noise elimination, outperforming competing probabilistic methods in this respect. Full topological flexibility was also demonstrated -- the algorithm correctly identifies non-linear structures, overcoming the sphericity limitation typical of purely centroid-based approaches.

\subsection{Verification of the Research Thesis}

Based on the evidence gathered, it can be confidently stated that the main objective of the paper has been fully achieved. The thesis put forward in the introduction -- that applying preliminary vector quantization makes it possible to reduce the computational complexity of density-based clustering while preserving the ability to detect non-spherical structures and robustness to noise -- has been unambiguously confirmed experimentally. It has been shown that reducing the input data to a compressed set of representatives is a highly effective and safe technique, provided that an appropriate approximation resolution is maintained, closely correlated with the size of the original dataset.

\subsection{Limitations of the Solution}

Despite the highly promising results, the studied algorithm exhibits certain limitations. The first is the complexity of parameterization, requiring the engineer to define both the number of micro-clusters and the density thresholds for the target structure. Although heuristics developed during the study facilitate the initialization process, improper calibration still carries the risk of masking topological structures or losing the performance gain. Another research challenge is the model's sensitivity to highly heterogeneous data density, which is inherited from the classical DBSCAN implementation \cite{ester1996density}. Analyzing highly non-uniform spaces would require the use of fully adaptive mechanisms. In addition, relying on stochastic optimization in the preliminary phase introduces a certain degree of non-determinism into the method, which, in systems with strict auditing requirements, forces the fixing of random-generator seeds or the averaging of results, affecting the final processing cost.

\subsection{Directions for Future Work and Vision Applications}

The development of the K-SCAN algorithm is aimed at increasing its production usefulness and opening entirely new application fields. An excellent example of its extension potential is a preliminary study of the method's application to digital image segmentation tasks. As part of an exploratory experiment, a portrait image of nearly 640 thousand pixels was segmented, aiming to extract the main semantic regions using 2000 intermediate nodes.

\begin{figure}[H]
    \centering
    \includegraphics[width=0.95\textwidth]{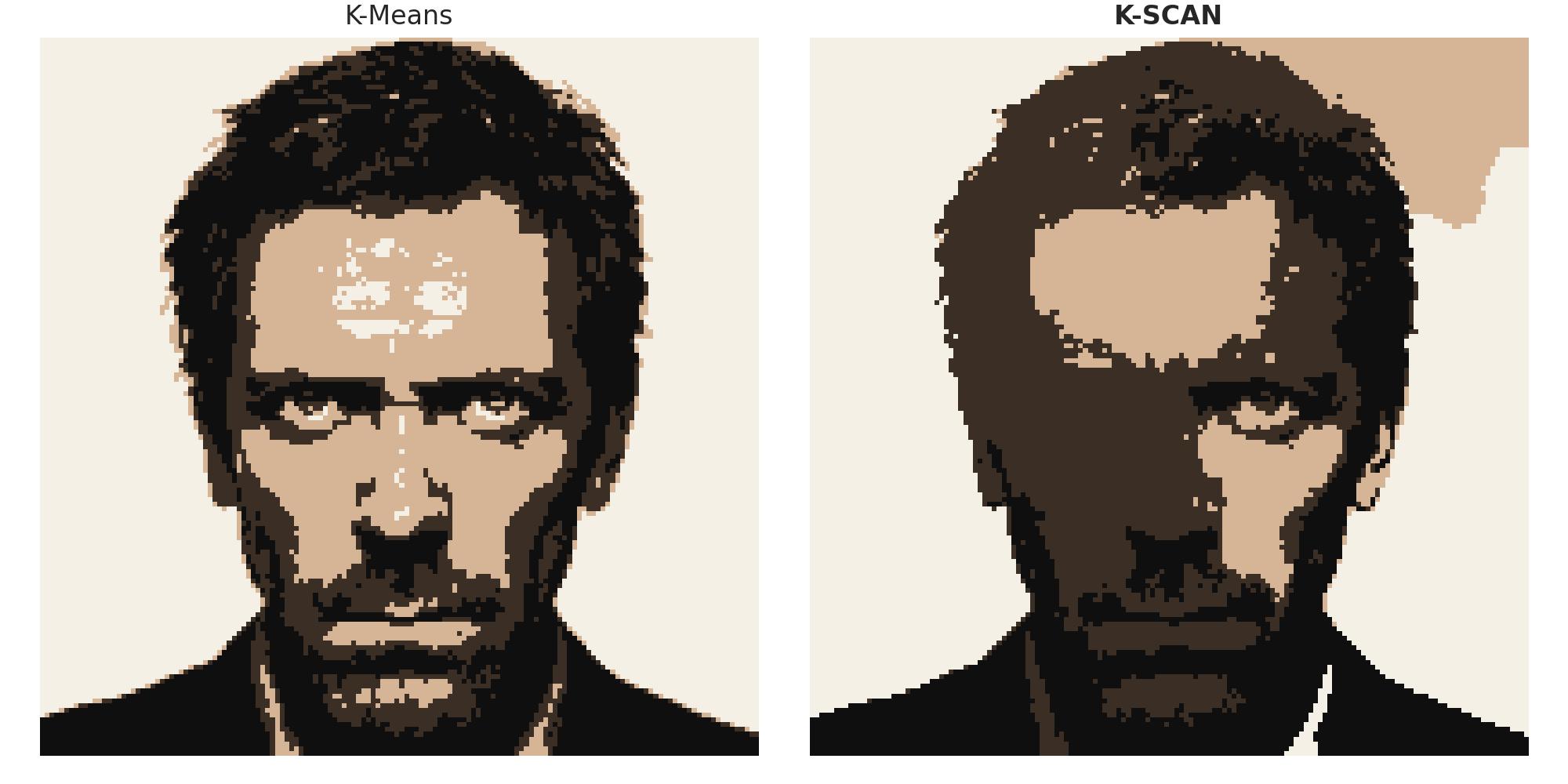}
    \caption{Comparison of segmentation paradigms. Left: K-Means (grainy image) groups pixels based solely on color. Right: K-SCAN (smooth-region effect) unifies the structure, eliminating background noise, but at the cost of losing the finest details.}
    \label{fig:house_comparison}
\end{figure}

The visual analysis of the compared paradigms, illustrated in Figure \ref{fig:house_comparison}, reveals the unique operating specifics of the hybrid. Thanks to the neighborhood analysis built into the mechanism, the method effectively eliminates spatial noise and creates uniform, smooth color patches, a property highly desirable, among other things, in lossy compression algorithms \cite{gersho1992vector} and artistic image stylization \cite{rosin2012image}. On the other hand, the model's absolute tendency to enforce topological consistency leads to a visible over-smoothing effect. As a result, small but perceptually crucial details -- represented by marginal statistical mass -- are filtered out as noise and absorbed into the dominant clusters.

Future optimization efforts in this area should focus on implementing adaptive sampling density based on local variance, as well as introducing an advanced feature vector that precisely balances the influence of color saturation and physical pixel distance. Beyond the exploration of visual spaces, the strictly technical development of the method is of key importance, including the development of parameter auto-tuning and the full porting of the computational logic to graphics-accelerator (GPU) architectures \cite{andrade2013g}. Parallelizing the compression stage would represent a breakthrough enabling the use of K-SCAN in a near-real-time mode.

Overall, this paper delivers a complete and verified analytical tool that combines the \textbf{high scalability} of partitional algorithms with the \textbf{noise robustness} characteristic of density-based methods. The developed solution can be applied in Business Intelligence systems and data engineering, where fast and precise exploration of large, noisy volumes of information is essential.

\vspace{1cm}
\section*{Computational Environment and Research Tools}
All numerical computations, data-processing procedures, and graphical visualizations presented in this paper were carried out using the Python programming language (version 3.12). The main algorithmic core and the evaluation of metrics were based on the optimized scientific libraries \texttt{scikit-learn} and \texttt{NumPy}. Plots and topological structures were generated using the \texttt{Matplotlib} library. Tests and computational simulations were carried out using the Google Colaboratory cloud environment.

\section*{Conflict of Interest Statement}
The authors declares no known competing financial interests or personal relationships that could have, or could appear to have, influenced the results of the research and the conclusions presented in this paper.

% --------------------------------------------------
% BIBLIOGRAPHY
% --------------------------------------------------
\cleardoublepage
\phantomsection
\addcontentsline{toc}{chapter}{Bibliography}
\nocite{*}
\bibliographystyle{abbrv}
\bibliography{bibliografia}

\end{document}